\documentclass{llncs}

\usepackage{makeidx}  

\usepackage{graphicx}
\usepackage{amsfonts}
\usepackage{cite}

\begin{document}


%
%
\title{The Active Atlas: Combining 3D Anatomical Models with Texture Detectors}
\author{Yuncong Chen\inst{1}, Lauren McElvain\inst{2}, Alex Tolpygo\inst{3}, Daniel Ferrante\inst{3}, \\Harvey Karten\inst{2}, Partha Mitra\inst{3}, David Kleinfeld\inst{2}, Yoav Freund\inst{1}}


%

\institute{Department of Computer Science and Engineering, University of California, San Diego, La Jolla, USA\\
\email{ \{yuncong, yoav\}@ucsd.edu}
\and Department of Physics, Univerity of California, San Diego, La Jolla, USA\\
\and Cold Spring Harbor Laboratory, Cold Spring Harbor, New York, USA\\}

\maketitle              
\begin{abstract}
While modern imaging technologies such as fMRI have opened exciting possibilities for studying the brain in vivo, histological
sections remain the best way to study brain anatomy at the level of neurons. 
The procedure for building histological atlas changed little since
1909 and identifying brain regions is a still a labor intensive process
performed only by experienced neuroanatomists. Existing digital atlases
such as the Allen Reference Atlas are constructed using downsampled images and can not reliably map low-contrast parts such as brainstem, which is usually annotated based on high-resolution cellular texture.

We have developed a digital atlas methodology that combines
information about the 3D organization and the detailed texture of different structures. Using the methodology we
developed an atlas for the mouse brainstem, a region for which there are currently no good atlases. Our atlas is ``active''
in that it can be used to automatically align a histological stack to
the atlas, thus reducing the work of the neuroanatomist.

\end{abstract}

\setcounter{footnote}{0}

\section{Introduction}

Pioneered by Korbinian Brodmannn in 1909\cite{brodmann2006brodmann}, the classical approach to mapping distinct brain regions is based on visually recognizing the cellular textures (cytoarchitecture) from images of sections of a brain. Several paper
atlases have been created in this way for the brains of different species \cite{paxinos2004mouse}. 

The primary methods for expert annotation of brain regions have changed little since then. It still is a labor intensive process performed only by the most experienced neuroanatomists.
In this paper we propose a machine learning approach for atlas construction that uses automated texture recognition to immitate human pattern recognition in the annotation task.


There exist several section-based digital atlases that were constructed using automated registration algorithms. The best known is the Allen Reference Atlas for mouse \cite{dong2008allen, ccf2015, fonov2011unbiased}, which is based on downsampled images of 50$\mu m$ per pixel. At this resolution, registration can be performed by maximizing intensity similarity using metrics such as correlation and mutual information.

\begin{figure}[t]
  \centering
    \includegraphics[width=1\textwidth]{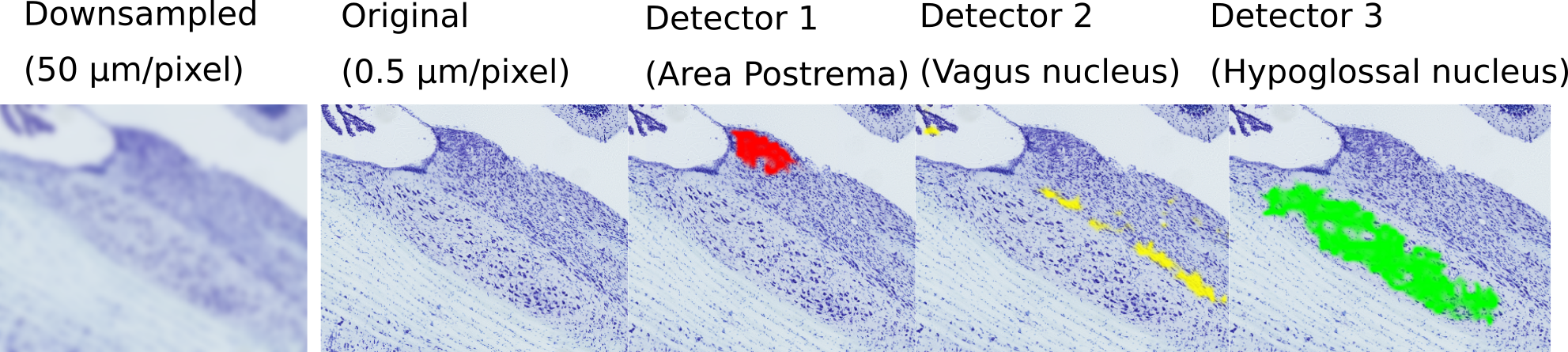}
  \caption{{\bf A demonstration of the limitation of reduced resolution
    brain images.} The ``Original'' image was taken at 0.5 $\mu m$/pixel. 
    ``Detector 1,2,3'' represent the detection of three brain structures based on texture by the trained classifiers. 
    The ``Downsampled'' image lacks the high-resolution details needed to distinguish the structure. (Best viewed in color)}
  \label{fig:grey_vs_texture}
\end{figure}

The problem is that at this resolution the information on cellular texture is discarded, which results in poor localization in regions that lack high contrast boundaries (see Figure~\ref{fig:grey_vs_texture}). 
In this work we focus on the mouse brainstem, a part that has numerous cytoarchitecturally identifiable nuclei but is relatively homogeneous at low resolution.
To overcome this limitation we have developed the {\em active atlas}, a texture-based atlas that operates on the full-resolution images and uses texture classifiers to differentiate structures not identifiable at low resolution. This distinguishes our approach from both the Allen atlas and those based on MRI or optical volumes \cite{johnson2010waxholm, mazziotta2001probabilistic, ronneberger2012vibe, peng2011brainaligner}.

The contributions of this work are:\\
\indent$\bullet$ Detection of cytoarchitectural textures visible only at high resolution.\\
\indent$\bullet$ Identification of discrete structures in addition to overall registration.\\
\indent$\bullet$ Characterization of the positional variability of brain structures.\\
\indent$\bullet$ Use of iterative refinement to reduce human annotation effort.

The paper is organized as follows. Section~\ref{sec:ActiveAtlas} describes the procedure for building an active atlas. Section~\ref{sec:Results} presents evaluation results that demonstrate the confidence of registration and accuracy of texture detection.

\section{The Active Atlas}
\label{sec:ActiveAtlas}

The active atlas has two components:
\begin{enumerate}
  \item {\bf Anatomical model:} stores for each of 28
    structures in the brainstem, the position
    statistics and probabilistic shape.
  \item {\bf Texture classifiers:} predict the probability
    that a given image patch corresponds to a particular structure.
\end{enumerate}

\begin{figure}[t]
  \centering
    \includegraphics[width=1\textwidth]{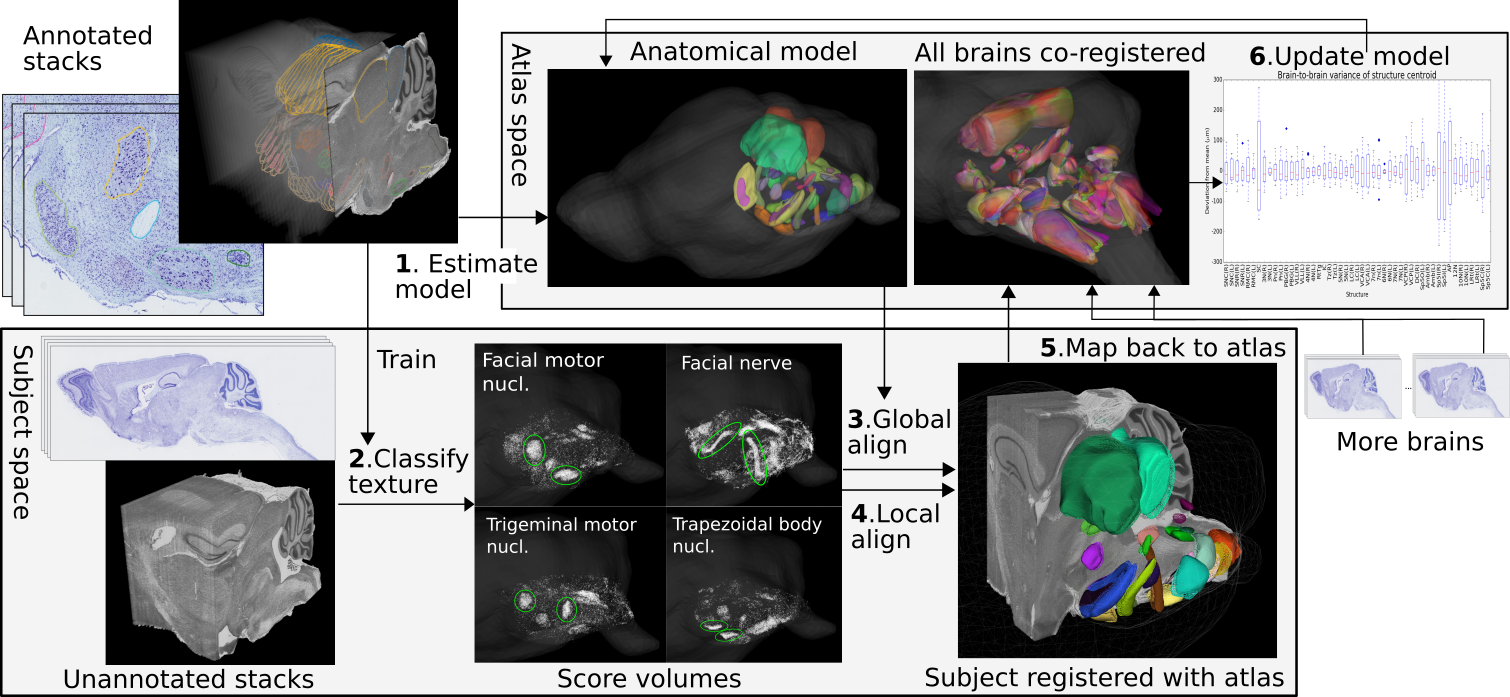}
  \caption{Incremental atlas building workflow}
  \label{fig:workflow}
\end{figure}

The construction of the atlas is iterative, starting with an
initialization step that required significant human labor, followed by refinement steps which require little or
no expert labor (see Figure~\ref{fig:workflow}).
In our case, the initial step was to annotate three stacks of images, which required 30 hours of work of an experienced neuroanatomist. 
From these annotated stacks, an initial anatomical model and a set of texture classifiers were constructed. 
The refinement then uses nine additional stacks that were {\em not} annotated. These stacks were aligned to the initial atlas and the information from this alignment was used to refine  the atlas and to estimate the variability from brain to brain. Below we provide more details on each step.

\noindent
    {\bf 2.1 Preprocessing.} Our dataset consist of sagittal brain sections from twelve mice of identical strain and age. 
    The 20$\mu m$ sections are mounted with a tape-transfer system \cite{pinskiy2015high} to ensure minimal distortion. 
    Each specimen gives roughly 400 sections, stained with Nissl and scanned at 0.5$\mu m$ resolution, demonstrating clear cytoarchitectonic features. 
    The sections of each brain are registered via in-plane correlation-maximizing rigid transforms and stacked to reconstruct a 3D volume. The sufficiency of rigid transforms is proved by the smooth structure boundaries on virtual coronal sections of reconstructed volumes.
  
\begin{figure}[t]
  \centering
    \includegraphics[width=1\textwidth]{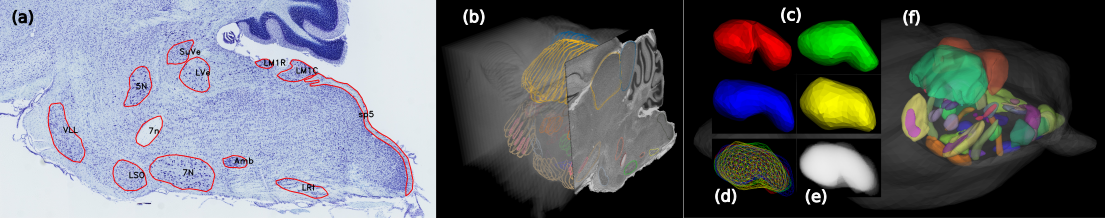}
  \caption{(a) Structure boundaries drawn by an expert (b) Aligned contour series in 3D (c) Facial motor nucleus from both hemispheres of different brains (d) Meshes aligned (e) Probabilistic average shape (f) Anatomical model with 28 structures}
  \label{fig:annotation}
\end{figure}

\noindent
    {\bf 2.2 Estimation of Anatomical Model.} Model estimation takes as input a
    current model (initially null) and a set of manually or automatically annotated brains.
From each annotated brain one can collect an aligned contour set for each structure (Figure~\ref{fig:annotation}b), which can be converted into a 3D mesh or volume (Figure~\ref{fig:annotation}c). Based on them we derive the average centroid position and the average shape of each structure, which constitute a refined model.

\noindent
    {\bf 2.2.1 Position Estimation.}
All brains are co-registered using the method described in Section 2.4. Centroid positions of the same structures in the common space are averaged over all brains. Those of paired structures are further adjusted to ensure symmetry of left and right hemispheres.
The covariance matrices of centroid positions are also computed. They quantify brain variability and are used as structure-specific constrains for aligning future data.

\newcommand{\vA}{\mathbf{A}}
\newcommand{\vR}{\mathbf{R}}
\newcommand{\vS}{\mathbf{S}}
\newcommand{\vL}{\mathbf{L}}
\newcommand{\vb}{\mathbf{b}}
\newcommand{\vp}{\mathbf{p}}
\newcommand{\vt}{\mathbf{t}}

\noindent
{\bf 2.2.2 Shape Estimation.} All meshes of the same structure are aligned using Iterative Closest Point algorithm\cite{besl1992method} (Figure~\ref{fig:annotation}d) and converted to aligned volumes. The average shape as a probabilistic volume is then computed by voxel-voting (Figure~\ref{fig:annotation}e).

Combining average shapes with average centroid positions, we obtain a probabilistic atlas volume $\vA$ where $\vA(\vp)$ denotes the 28-dimensional probability vector at location $\vp$.

\noindent
{\bf 2.3 Learning Texture Classifiers.} We train texture classifiers to differentiate a structure from its immediate surrounding region. 
We found that this gives better results than training against the entire background. 
The probable reason is that the anatomical model eliminates most of the uncertainty in gross positions, allowing the texture classifiers to focus on correcting small-scale error. 

Image patches roughly 100$\mu m$ by 100$\mu m$ are used as units for classification. For each
structure, a binary logistic regression classifier is trained using a
positive patch set extracted from the interior of structure
boundaries and a negative set extracted from the surrounding region
within 50$\mu m$ from the boundaries.
%
The feature vectors encoding the patches are the 1024-dimensional output of a pre-trained deep convolutional neural network (Inception-BN\cite{ioffe2015batch}). 
Although the network was originally trained for classifying natural images, it proves
effective also for classifying histology textures.

\begin{figure}[t]
  \centering
    \includegraphics[width=1\textwidth]{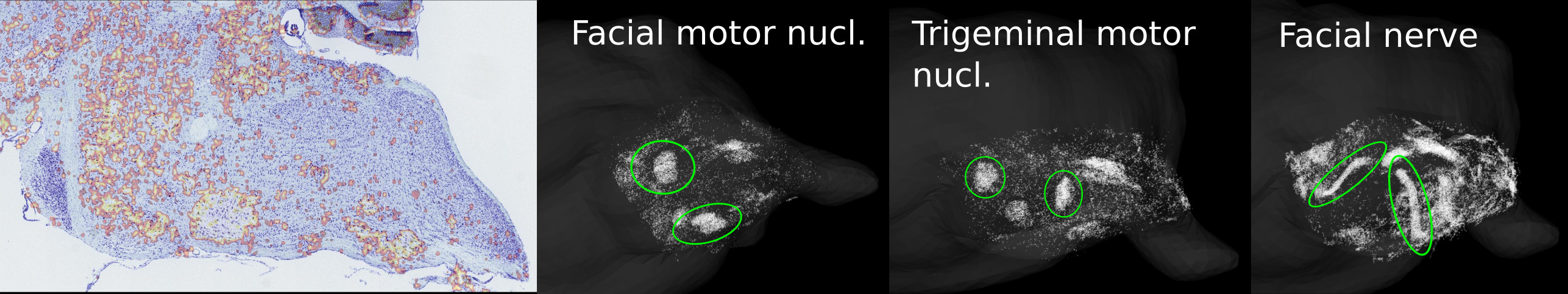}
  \caption{(a) An example score map for facial motor nucleus (b) Stacking 2D score maps forms a 3D score volume. (c,d) Score volumes for other structures.}
  \label{fig:learning}
\end{figure}

For an unannotated image, these classifiers are applied to patches with 25$\mu m$ spacing, 
resulting in score maps for different structures.
All score maps of a same structure in one stack undergo the previously computed intra-stack alignment to form a set of 3D score volumes. 
Each volume represents a probabilistic estimate of a particular structure's position in the reconstructed specimen (Figure \ref{fig:learning}). 
Denote by $\vS(\vp)$ the vector consisting of the scores for different structures at location $\vp$.

\noindent
{\bf 2.4 Registering Atlas to Specimen.}
Registration is driven by maximizing the correlation at all voxels between the
score vectors of the specimen volume and the probability vectors of the atlas volume. 
A global 3D transform first aligns the atlas roughly with the whole specimen. 
Affine transform is used to account for non-vertical cutting angle and scale change due to
dehydration. 
Separate 3D translations are then applied to each structure so independent variations can be captured. 

Let $\Omega$ be the domain of the atlas.
For global transform, the objective to maximize is simply $F^{global}(\vL, \vb) = \sum_{\vp \in \Omega} \vA(\vp) \cdot \vS(\vL\vp + \vb)$, where $\vL\in \mathbb{R}^{3\times3}$ and $\vb\in\mathbb{R}^3$ are respectively the linear and translation parts of the affine transform.

For the local transform of structure $k$, only the voxels inside the structure and those in a surrounding region within a 50$\mu m$ radius are concerned. Denote the two sets by $\Omega_k^+$ and $\Omega_k^-$ respectively. The objective is
\begin{equation}
F^{local}(\vt) = \sum_{\vp\in \Omega_k^+} \vA(\vp) \cdot \vS'(\vp+\vt) - \sum_{\vp\in \Omega_k^-} \vA(\vp) \cdot \vS'(\vp+\vt)\; - \; \eta
\vt^\mathsf{T}C_k\vt \;,
\end{equation}
where $\vt \in \mathbb{R}^3$ is the local translation and $\vS'$ is the globally transformed score volume.
The regularization term penalizes deviation from the mean position defined in the atlas model, where $C_k$ is the inverse of the position covariance matrix (see Section 2.2.1).

\begin{figure}
  \centering
    \includegraphics[width=1\textwidth]{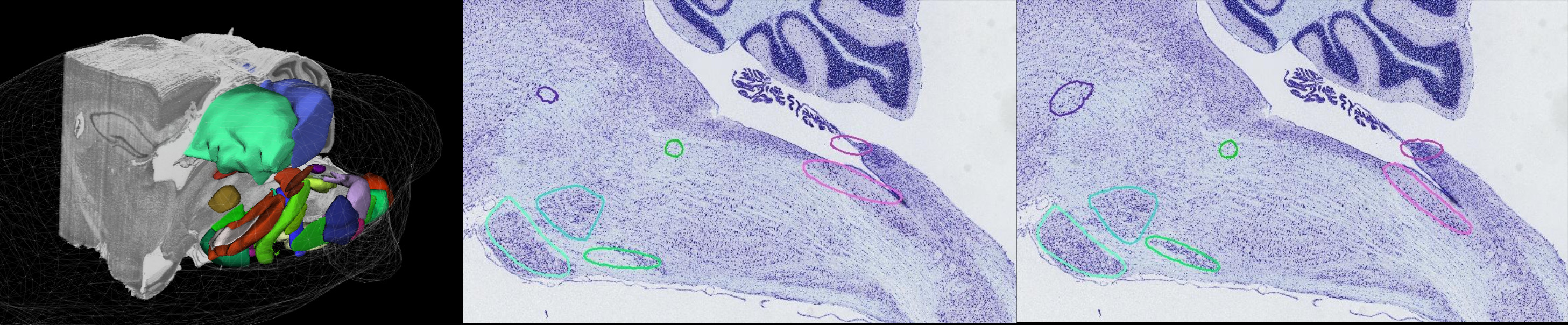}
  \caption{(a) Reference model globally registered to the specimen. (b) Global
    registration. Showing the structure contours on a
    section. Structures are roughly aligned. (c) Local
    registration. Structures are aligned perfectly.}
  \label{fig:registration}
\end{figure}

Optimization for both cases starts with grid search, followed by gradient
descent where the learning rate is determined using Adagrad\cite{duchi2011adaptive}. From Figure \ref{fig:registration} one can visually verify the accuracy of registration. This registration effectively annotates new stacks for the 28 structures.

\begin{figure}[t]
  \centering
    \includegraphics[width=1\textwidth]{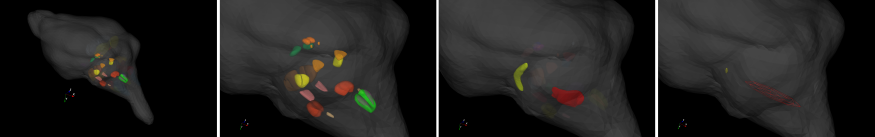}
  \caption{(a, b) Confident structures (c) Two unconfident structures (d) Uncertainty ellipsoids. The elongated structure VLL (yellow) is uncertain only in its axial direction, while Sp5I (red) is uncertain in rostral-caudal direction because its rostral and caudal boundaries are ambiguous. (Best viewed in color)}
  \label{fig:3d}
\end{figure}

\noindent
{\bf 2.5 Evaluating Registration Confidence.}
The registration algorithm seeks a local maxima of the objective functions. 
We quantify the confidence of the registration by considering the height and
the width of the converged local maxima. The height
of the peak is normalized by considering a $z$-test relative to the
variance within a sphere around the peak. The width can be computed for any direction, based on the Hessian of the $z$-score around the peak, as the distance away from peak that the $z$-score drops to 0. 
Figure \ref{fig:3d} shows examples where different directions have different localization confidence.


%


\noindent
{\bf 2.6 Updating Atlas.}
After new brains are co-registered with the atlas, average positions and shapes for all structures are re-estimated. Additional training patches can also be collected from the automatic annotations to improve the classifiers.

\section{Results}
\label{sec:Results}

\begin{figure}[t]
  \centering
    \includegraphics[width=1\textwidth]{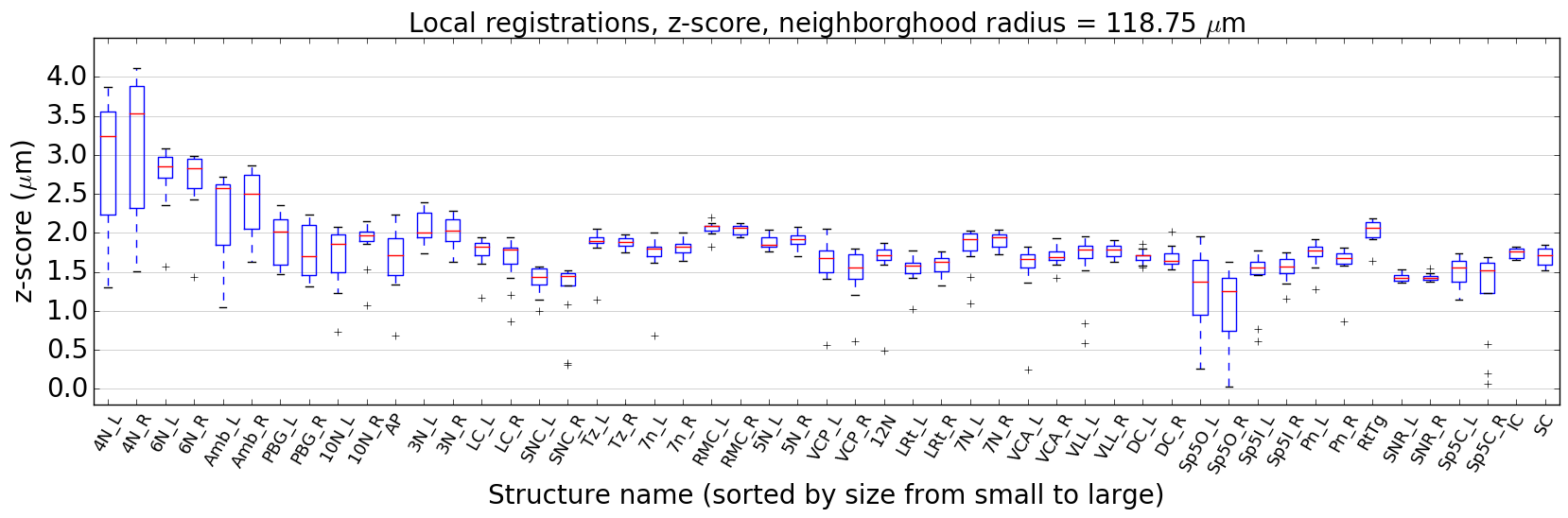}
  \caption{Average z-scores of the local registrations of different structures.}
  \label{fig:Confidence}
\end{figure}

\begin{figure}[t]
  \centering
    \includegraphics[width=1\textwidth, height=180pt]{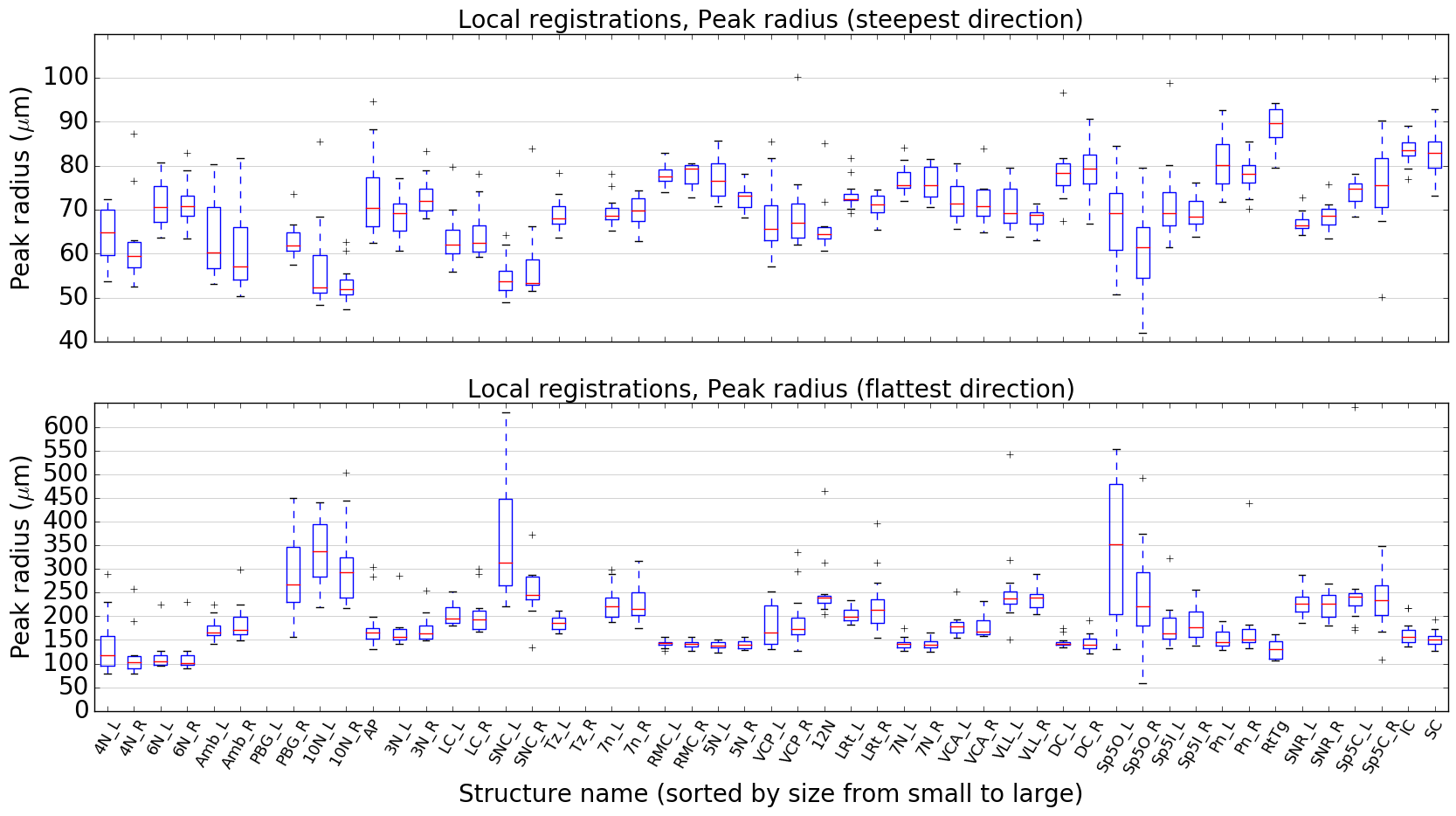}
  \caption{Average peak width of per-structure registrations of different structures.}
  \label{fig:peak_radius}
\end{figure}

\noindent
{\bf 3.1 Confidence of Registrations.}
The global registrations across all specimens have an average peak z-score of 2.06. The average peak radius is 98$\mu m$ in the steepest direction and 123 $\mu m$ in the flattest direction. This suggests that the derived reference model captures the common anatomy of this population and matches all specimens with little space for adjustment. Figure \ref{fig:Confidence} and \ref{fig:peak_radius} show these for the per-structure registrations. 
The average z-score is 1.79 and the width is between 90$\mu m$ and 250 $\mu m$ for most structures. Generally, small structures tend to be registered more confidently than large ones. This aligns well with intuition if one considers how position shifts affect the overlap between the structure and the texture map. For a small structure, a small translation might completely eliminate any overlap, while a large structure is less sensitive.

\begin{figure}[t]
  \centering
    \includegraphics[width=1\textwidth]{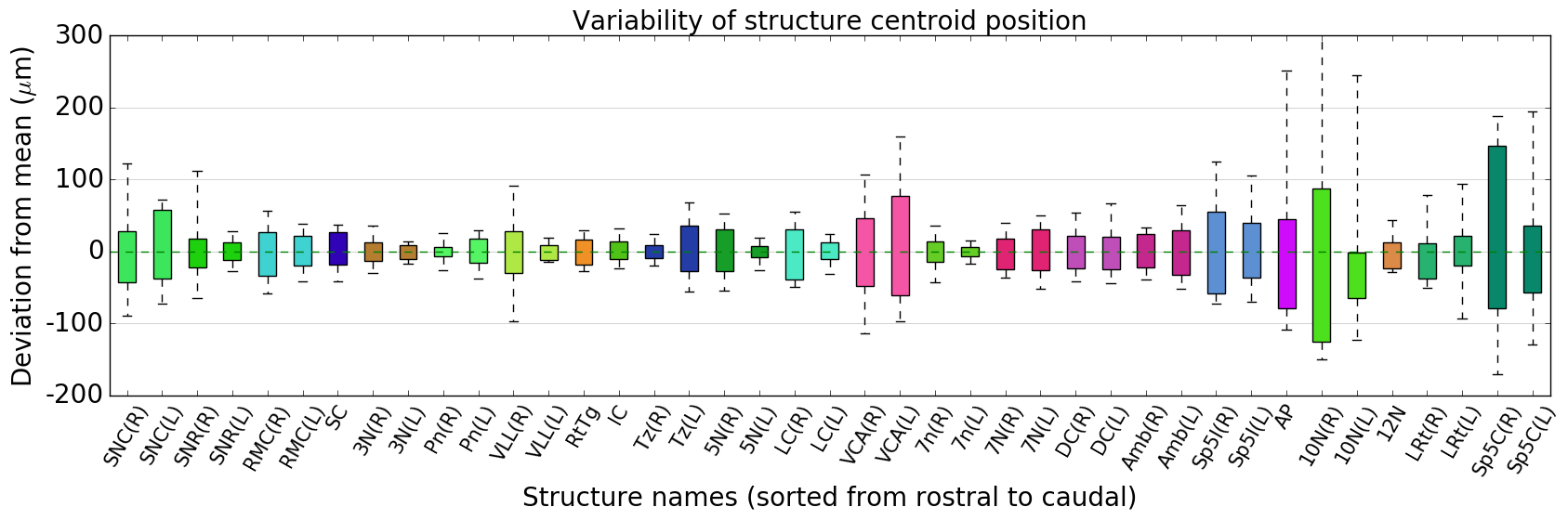}
  \caption{Variability of centroid positions for different structures. Same color indicates the same structure in left (L) and right (R) hemispheres.}
  \label{fig:variation}
\end{figure}

\noindent
{\bf 3.2 Variability of Structure Position.}
Variability is captured by the amount of per-structure translation. 
Figure \ref{fig:variation} shows these for different structures across all specimens. 
Most structures vary within 100um of the mean position defined in atlas. Some structures are particularly variable, which are also the ones whose boundaries are difficult to define. The same structure in left and right hemispheres generally have similar variability.

%



\begin{figure}
  \centering
    \includegraphics[width=1\textwidth]{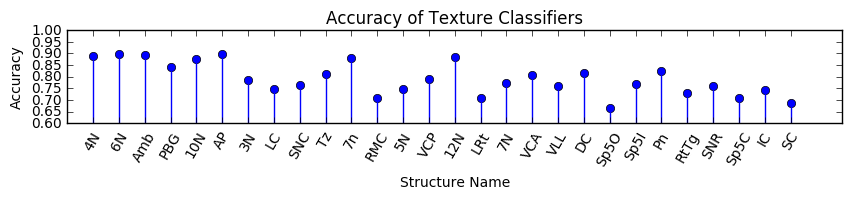}
  \caption{Accuracy of texture classifiers. (Structures sorted by increasing size)}
  \label{fig:classifier_accuracy}
\end{figure}

\noindent
{\bf 3.3 Accuracy of Texture Classifiers.}
Figure \ref{fig:classifier_accuracy} shows the test accuracy for the classification of different structures. They range from 0.7 to 0.9 with a mean of 0.79. Larger structures tend to be harder to classify possibly due to their texture being more inhomogeneus.


\section{Conclusion}

The results demonstrate a form of co-training between the anatomical model and the texture classifiers. On the one hand, registrations perform well despite the classifiers for some structures are suboptimal, due to the strong constraint by the anatomical model. On the other hand, confident detection of the characteristic textures of many structures allows specimen-specific deviations from the current anatomical model to be discovered, contributing to more accurate variability. The synergy between the anatomical information and texural information is the key feature of the proposed active atlas.

\bibliography{miccai_arxiv}
\bibliographystyle{splncs03}

\end{document}